%% file: main.tex
\pdfoutput=1
\documentclass[11pt]{article}
\usepackage{acl}

\usepackage{amsmath}
\usepackage{balance}
\usepackage{times}
\usepackage{latexsym}
\renewcommand{\UrlFont}{\ttfamily\small}

\newcommand{\added}[1]{#1}
\newcommand{\was}[1]{}

\usepackage{microtype}

\def\aclpaperid{10} 

\usepackage{pdfpages}

\title{Conversational Search with Mixed-Initiative - Asking Good Clarification Questions backed-up by Passage Retrieval}

\author{Yosi Mass, Doron Cohen, Asaf Yehudai
and David Konopnicki\thanks{work done while at IBM} \\
  IBM Research AI\\
  Haifa University, Mount Carmel, Haifa, HA 31905, Israel \\
  {\tt \{yosimass,doronc\}}@il.ibm.com\\
  {\tt {Asaf.Yehudai}}@ibm.com\\
  {\tt {dkonopnicki}}@gmail.com
  }
\date{}

\begin{document}
\maketitle
\input{abstract}
\input{introduction}
\input{related}

\input{method}

\input{experiments}

\input{conclusions}

\bibliography{anthology,references}

\appendix
\section{Appendix}
\input{Appendix/passage_retrieval}
\end{document}

%% file: abstract.tex
\begin{abstract}
We deal with the scenario of conversational search, where user queries are under-specified or ambiguous. This calls for a mixed-initiative setup.  User-asks (queries) and system-answers, as well as system-asks (clarification questions) and user response, in order to clarify her information needs. 
We focus on the task of selecting the next clarification question, given the conversation context. Our method leverages passage retrieval from a background content 
\was{that is used both for an initial selection of relevant candidate clarification questions, as well as} to fine-tune two deep-learning models for ranking candidate clarification questions. We evaluated our method on two different use-cases. The first is an open domain conversational search in a large web collection. The second is a task-oriented customer-support setup.
We show that our method performs well on both use-cases.

\end{abstract}

%% file: introduction.tex
\section{Introduction}
\label{sec:introduction}
A key task in information and knowledge discovery is the retrieval of relevant information given the user's information need (usually expressed by a query). With the abundance of textual knowledge sources and their diversity, it becomes more and more difficult for users, even expert ones, to query such sources and obtain valuable insights. 

Thus, users need to go beyond the traditional ad-hoc (one-shot) retrieval paradigm. This requires to support the new paradigm of conversational search – a sophisticated combination of various mechanisms for exploratory search, interactive IR, and response generation. In particular, the conversational paradigm can support mixed-initiative: namely, the traditional user asks - system answers interaction in addition to system-asks (clarification questions) and user-answers, to better guide the system and reach the information needed~\cite{Krasakis_2020}.


Existing approaches for asking clarification questions include \textit{selection} or \textit{generation}. In the selection approach, the system selects clarification questions from a pool of pre-determined questions~\cite{Aliannejadi-sigir19}.
In the generation approach, the system generates clarification questions using rules or using neural 
generative 
models~\cite{zamani20}.

In this work we focus on the selection task.
\added{While the latter (i.e., generation) may represent a more realistic use-case, still there is an interest in the former (i.e., selection) as evident by the Clarifying Questions for Open-Domain Dialogue Systems (ClariQ) challenge ~\cite{aliannejadi2020convai3}. Moreover, the selection task represents a controlled and less noisy scenario, where the pool of clarifications can be mined from e.g., query logs.}

In this paper we deal with content-grounded conversations. Thus, a conversation starts with an initial user query, continues with several rounds of conversation utterances (0 or more), and finally ends with one or more documents being returned to the user.  
Some of the agent utterances are marked as clarification questions. 

The task at hand is defined as follows. Given a conversation context up to (and not including) a clarification-question utterance, predict the next clarification question. A more formal definition is given in Section~\ref{sec:method} below.

\begin{figure}[tbh]
    \centering
    \includegraphics[width=3in]{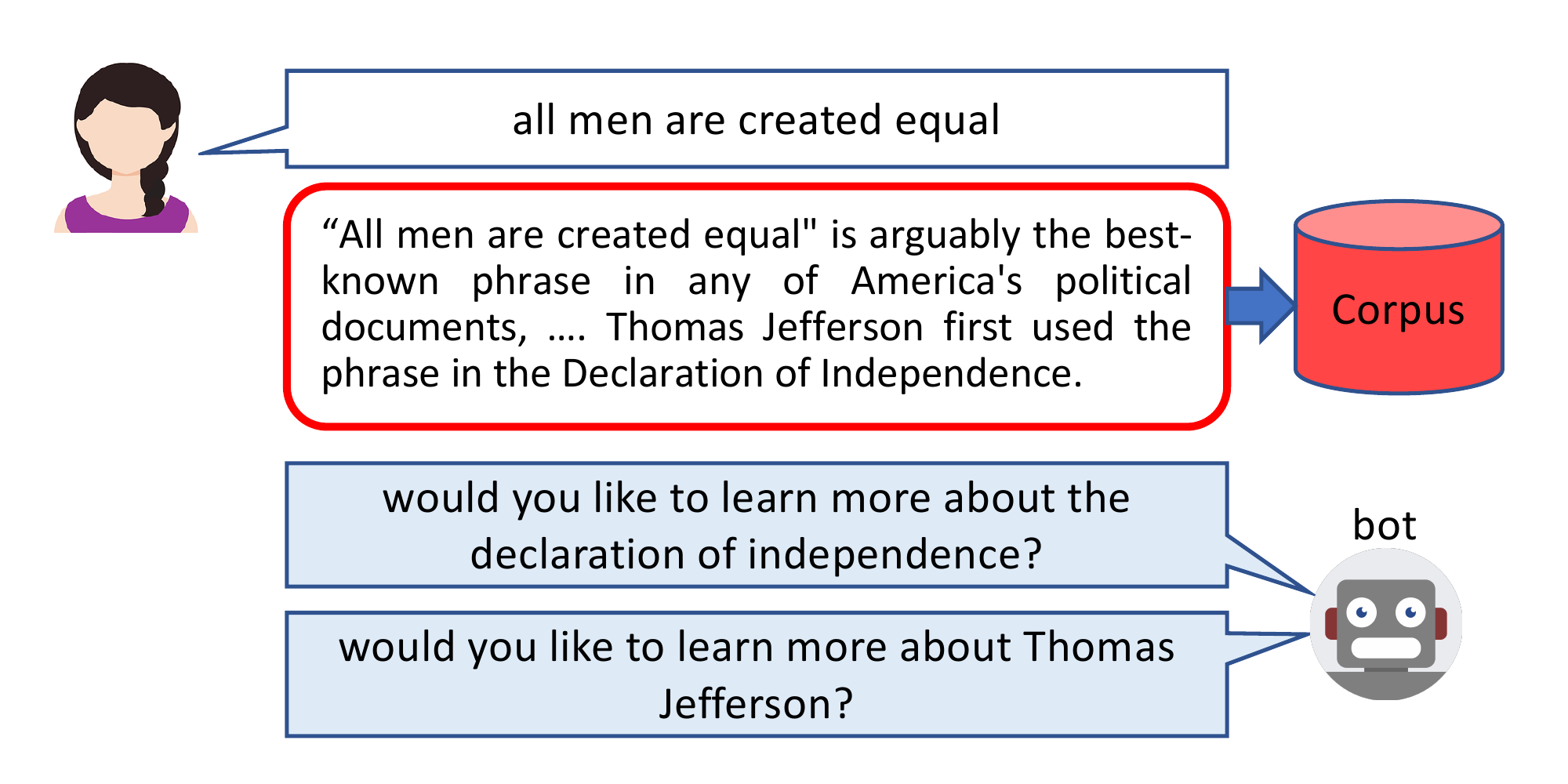}
    \caption{A motivating example}\vspace{-.5cm}
    \label{fig:example}
\end{figure}

Intuitively, clarification questions should be used to distinguish between several possible intents of the user. We approximate those possible intents through passages that are retrieved from a given corpus of documents.
\was{The system is trained to select clarification questions, leveraging potential relevant passages that can answer the user information need.} 
\was{Passage retrieval is used as a mediator both for an initial retrieval of candidate clarification questions, as well as for training deep learning models for re-ranking these clarification questions.} \added{A motivating example from the ~\cite{aliannejadi2020convai3} challenge is given in Figure~\ref{fig:example}. The user wants to get information about the topic \textit{all men are created equal}. Through the retrieved passage, the system can ask the mentioned clarification questions. 
}

We use two deep-learning models. The first one learns an association between conversation context and clarification questions. The second learns an association between conversation context, candidate passages and clarification questions. 

Evaluation was done on two different use-cases. The first one is an open domain search in a large web corpus~\cite{aliannejadi2020convai3}. The second is an internal task-oriented customer-support setup, where users ask technical questions. We show that our method performs well on both use-cases. 




%% file: related.tex
\section{Related work}
\label{sec:related}


We focus on works that deal with clarification-questions selection. 
\citet{Aliannejadi-sigir19} describes a setup very similar to ours for the aforementioned task.
They apply a two-step process. In the first step, they use BERT~\cite{BERT} to retrieve candidate clarification questions and, in the second step, they re-rank the candidates using multiple sources of information. Among them are the scores of retrieved documents using the clarification questions. 
However, they do not look at passage content as we do. 

The ClariQ\footnote{http://convai.io} challenge organized a competition for selecting the best clarification questions in an open-domain conversational search. The system by NTES\_ALONG  ~\cite{NTES-ALONG} was ranked first. They first retrieve candidate clarification questions and then re-rank them using a ROBERTA~\cite{Roberta} model, that is fine-tuned on the relation between a query and a clarification question.
Unlike our method, they do not exploit passage content. 

In~\citet{rao-daume-acl-2018}, they select clarification questions using the expected value of perfect information, namely a good question is one whose expected answer will be useful. 
They do not assume a background corpus of documents.


%% file: method.tex
\section{Clarification-questions Selection}

\subsection{Problem definition}
\label{sec:problem_definition}
A conversation $C$ is a list of utterances, $C=\{c_0, ..., c_n\}$ where $c_0$ is the initial user query. Each utterance has a speaker which is either a user or an agent.\footnote{An agent can be either a human agent or a bot.} Since we deal with content-grounded conversations, the last utterance is an agent utterance, that points to a document.  

We further assume that agent utterances are tagged with a \textit{clarification flag} where a value of $1$ indicates that the utterance is a clarification question. This flag is either given as part of the dataset (e.g., in the open domain dataset, ClariQ) or is derived automatically by using a rule-based model 
or a classifier. We discuss such rules for the second task-oriented customer-support dataset (see Section~\ref{sec:datasets} below).  

The \textbf{Clarification-questions Selection} task is defined as follows. Given a conversation context $C^{j}=\{c_0,...,c_{j-1}\}$, predict a clarification question at the next utterance of the conversation.\footnote{We always return clarification questions. We leave it for future work to decide whether a clarification is required.} 

\subsection{Method}
\label{subsec:method}
\begin{figure*}[tbh]
    \centering
    \includegraphics[width=6in]{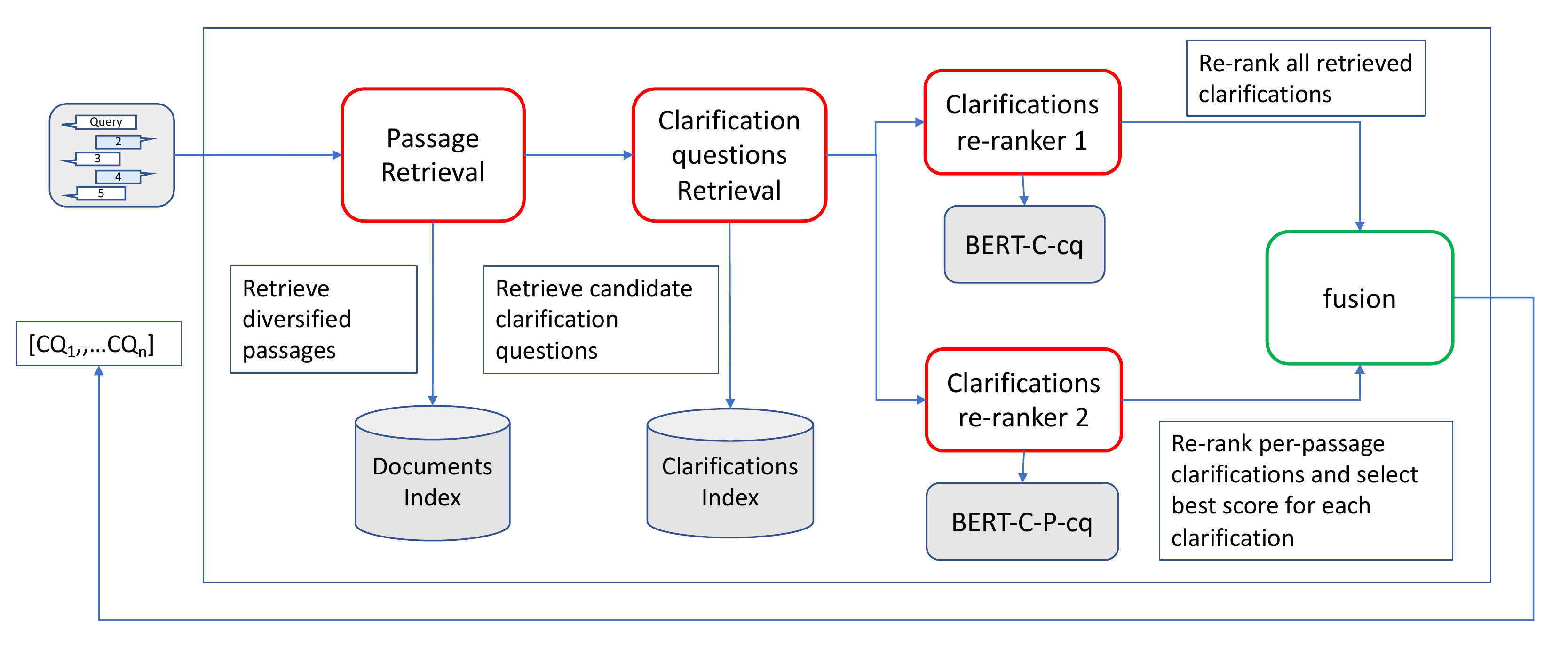}
    \caption{Clarification-questions selection run-time architecture}\vspace{-.5cm}
    \label{fig:arch}
\end{figure*}

The proposed run-time architecture is depicted in Figure~\ref{fig:arch}.
It contains two indices and two fine-tuned BERT models. The  \textit{Documents index} contains the corpus of documents (recall that we deal with conversations that end with a document(s) being retrieved). This index supports passage retrieval. 
The \textit{Clarification-questions index} contains the pool of clarification questions. The two BERT models are used for re-ranking of candidate clarification questions as described below.  

\label{sec:method}

Given a conversation context $C^j$, we first retrieve top-k passages from the Document index (See Section~\ref{sec:passage} below). 
We then use 
those passages, to retrieve candidate clarification questions from the Clarification-questions index  (See Section~\ref{sec:retrieval} below).
We thus have, for each passage, a list of candidate clarification questions.

The next step re-ranks those candidate clarification questions.
Re-ranking is done by the fusion of ranking obtain through two BERT models. 
Each model re-ranks the clarification questions by their relevance to the given conversation context and the retrieved passages (see Section~\ref{sec:re-ranking} below).
The components of the architecture are described next in more details.

\input{passage-retrieval.tex}

\input{clarification-retrieval.tex}
\input{clarification-reranking}

%% file: passage-retrieval.tex
\subsection{Conversation-based passage retrieval}
\label{sec:passage}

Documents in the document index are represented using three fields. The first field contains the actual document content. The second field augments the document's representation with the text of all conversations that link to it in the train-set~\cite{amitayQanchors}. We also keep a third field that contains the concatenation of the above two fields. We refer to those fields as \texttt{text}, and \texttt{anchor} and \texttt{anchor\_and\_text}  respectively. 

Given a conversation context $C^j$, Passage retrieval is performed in two steps. First, top-k documents are retrieved using those fields (we discuss in Section~\ref{sec:datasets} below, how to decide which field to use for retrieval). 
Following~\cite{agent-assist}, we
treat the conversation as a verbose query 
and create a disjunctive query over all words in the conversation $C^j$. 
We then apply the Fixed-Point (FP) method~\cite{Paik2014} for weighting the query words. Yet, compared to ``traditional'' verbose queries, conversations are further segmented into distinct utterances. Using this observation, we implement an \textit{utterance-biased} extension for enhanced word-weighting. To this end, we first score the various utterances based on the initial FP weights of words they contain. We then propagate utterance scores back to their associated words.

In the second step, candidate passages are extracted from those top-k documents using a sliding window of fixed size with some overlap. 
Each candidate passage $p$ is assigned an initial score based on the coverage of terms in $C^j$ by $p$. The coverage is defined as the sum over all terms in each utterance, using terms' global~ $idf$ (inverse document frequency) and their (scaled) $tf$ (term frequency). The final passage score is a linear combinations of its initial score and the score of the document it is extracted from.
Details are given in appendix~\ref{appendix:passage}



%% file: clarification-retrieval.tex
\subsection{Clarification-questions retrieval}
\label{sec:retrieval}

The pool of clarification questions is indexed into a Clarification index.
We use the passages returned for a given conversation context $C^j$, 
to extract an initial set of candidate clarification questions as follows.
For each passage $P$, we concatenate its content to the text of all utterances in $C^j$, and use it as a query to the Clarification index.

We thus have, for each passage, a list of candidate clarification questions. 



%% file: clarification-reranking.tex
\subsection{Clarification-questions re-ranking}
\label{sec:re-ranking}

The input to this step is a conversation context $C^j$, a list of candidate passages, and a list of candidate clarification questions for each passage. 
We use two BERT~\cite{BERT} models to re-rank the candidate clarification questions. The first model, \textit{BERT-C-cq} learns an association between conversation contexts and clarification questions. The second model, \textit{BERT-C-P-cq} learns an association between conversation contexts, passages and clarification questions. Training and using the two models is described below. 

\noindent \textbf{Fine-tuning of the models.} 
The first model, BERT-C-cq, is fine-tuned through a triplet network~\cite{hoffer_ailon2015} that is adopted for BERT fine-tuning~\cite{mass2019study}. It uses triplets $(C^j, cq^{+}, cq^{-})$, where $cq^{+}$ is the clarification question of conversation $C$ at utterance $c_j$ (as given in the conversations of the training set). Negative examples ($cq^{-}$) are randomly selected from the pool of clarification questions (not associated with $C$). 


For fine-tuning the second model, BERT-C-P-cq, we need to retrieve relevant passages. 
We use a weak-supervision assumption that all passages in a relevant document (i.e., a document returned for $C$), are relevant as well.
\was{Given a conversation $C$ in the training set, with $c_j, c_k \in C$, a clarification question and its answer utterances  respectively ($k \ge j$),
we retrieve passages using the conversation context $C^{k+1}$ (namely we take all the first utterances, including the clarification question and its answer), 
while limiting the retrieved passages to come from the relevant documents of $C$ only. 

To have high quality weak-supervised passages we use for training, only clarification questions whose answer is positive (i.e., contains positive words such as \textit{Yes} and no negative words).
} A triplet for the second BERT model is thus $(C^j \hspace{2pt}[SEP]\hspace{2pt}P, cq^{+}, cq^{-})$, where $P$ is a passage retrieved for $C^j$, $[SEP]$ is BERT's separator token, $cq^{+}$ and $cq^{-}$ are positive and negative clarification questions selected as described above for the first model.
\was{is a clarification question (with its answer utterance $c_k$) in $C$, and P is a passage retrieved using the conversation context $C^{k+1}$. }
\was{Negative examples $cq^{-}$ are selected similar to the first model.}

Due to the BERT limitation on max number of tokens (512), we represent a conversation context $C^j$ using the \was{initial user query, concatenated to the last $m$ utterances}
last $m$ utterances 
whose total length is less than 512 characters. 
We also take the passage window size to be 512 characters.\footnote{note that BERT uses tokens while for the passages and representation of conversation we use characters}

\was{\noindent\textbf{Re-ranking with the models.} Each candidate clarification question $cq_i$ is fed to the first model with the conversation context, namely $(C^j,cq_i)$.  Let $P$ be the passage that was used to retrieve $cq_i$. We feed $(C^j \hspace{2pt}[SEP]\hspace{2pt}P, cq_i)$ to the second model. Final scores of the candidates is set by simple CombSUM~\cite{combsum} fusion of their scores from the two BERT models.}
\noindent\textbf{Re-ranking with the models.} Each candidate clarification question $cq_i$ is fed to the first model with the conversation context as $(C^j,cq_i)$, and to the second model as $(C^j \hspace{2pt}[SEP]\hspace{2pt}P, cq_i)$, where $P$ is the passage that was used to retrieve $cq_i$. Final scores of the candidates is set by simple CombSUM~\cite{combsum} fusion of their scores from the two BERT models.


%% file: experiments.tex
\section{Experiments}
\label{sec:exp}

\input{experiments-datasets}
\input{experiments-setup}
\input{experiments-results}

%% file: experiments-datasets.tex
\subsection{Datasets}
\label{sec:datasets}

We evaluated our method on two datasets.
The first, \textbf{ClariQ}~\cite{aliannejadi2020convai3} represents an information-seeking use-case. 
The second, \textbf{Support} contains conversations and technical documents of an internal customer support site. Statistics on the two datasets are given in Table~\ref{tab:datasets}.

\was{There are several differences between the two datasets. First, }
The \textbf{ClariQ} dataset was built by crowd sourcing for the task of clarification-questions selection, thus it has high quality clarification questions. Each conversation has exactly three turns. Initial user query, an agent clarification question and the user response to the clarification question. The agent utterance is always a clarification question.  

The \textbf{Support} dataset contains noisy logs of human-to-human conversations, that contain a lot of chit-chat utterances such as 
\textit{Thanks for your help} or 
\textit{Are you still there?}
We thus applied the following rules to identify agent clarification questions. i) We consider only sentences in agent utterances that contain a question mark. ii) We look for question words in the text (e.g., \textit{what, how, where, did, etc.}) and consider only the text between such a word and the question mark. iii) If no question words were found, we run the sentences with the question mark through Allennlp's constituency parser~\cite{joshi2018extending}, and keep sentences with a Penn-Treebank clause type of \textit{SQ} or \textit{SBARQ}\footnote{\UrlFont{https://gist.github.com/nlothian/9240750}}. 

The above rules can be used to detect question-type sentences. However, we are interested in clarification questions that are related to the background collection of documents and not in chit-chat questions (such as e.g., \textit{how are you today?}). To filter out such chit-chat question types, we apply a 4th rule as follows. iv) Recall that each conversation ends with a document answer. We send the detected question and its answer (the next user's utterance), as a passage retrieval query (see Section 3.1 above) to the Documents index and keep only those questions that returned in their top-3 results, a passage from the document of the conversation.   

\was{
Second, in \textbf{ClariQ} all utterances are clarification questions and their answers. There is no real conversation context as in  the \textbf{Support} dataset. In other words, each topic (i.e., user basic question) in \textbf{ClariQ} has independent set of clarification questions and their answers. }

\begin{table}[tbh]
\small
\caption{Datasets statistics}
	\label{tab:datasets}
	\center
    \begin{tabular}{|l|c|c|}
        \hline
        & ClariQ & Support \\ \hline
        \#docs & 2.7M & 520 \\ \hline
        \#conversations (train/dev/test) & 187/50/60 & 500/39/43 \\ \hline
        \#total clarifications & 3940 & 704 \\ \hline
        \#avg/max turns per C & 3/3 & 8.2/80.5 \\ \hline
        \#avg/max clarifications per C & 14/18 & 1.27/5 \\ \hline
    \end{tabular}\vspace{-.5cm}
\end{table}

%% file: experiments-setup.tex
\subsection{Setup of the experiments}
\label{sec:setup}

We use Apache 
Lucene\footnote{https://lucene.apache.org/} 
for indexing the documents. We use English language analyzer and default BM25 similarity~\cite{RobertsonBM25}. 

For the customer support dataset (\textbf{Support}) we used the \texttt{anchor\_and\_text} field for initial document retrieval, since most documents in the dataset do have training conversations. 
\was{since it has quite few documents (only 520) and there is a large overlap between documents in the train and dev/test conversations.} 

The open-domain dataset (\textbf{ClariQ}) contains a large number of documents (2.7M), but only a small portion of them do have training conversations. Using the \texttt{anchor\_and\_text} field for retrieval will prefer that small subset of documents (since only they have anchor text). Thus for this dataset, we used the \texttt{text} field for retrieval. 

For passage retrieval, we used a sliding window of $512$ characters on retrieved documents' content. 
We used common values for the hyper parameters, with $\lambda=0.5$ to combine document and passage scores, and $\mu=2000$ for the dirichlet smoothing of the documents LM used in the FixedPoint re-ranking. Details of the passage retrieval are given in Apendix~\ref{appendix:passage}. 

The full conversations were used to retrieve passages. For feeding to the BERT models, we concatenated the last $m$ utterances whose total length was less than $512$ characters (we take full utterances that fit the above size. We do not cut utterances).

We used the pytorch huggingface implementation of BERT\footnote{\url{https://bit.ly/2Me0Gk1}}. For the two BERT models we used bert-base-uncased ($12$-layers, $768$-hidden, $12$-heads, $110$M parameters). Fine-tuning was done with the following default hyper parameters.   max\_seq\_len of 256 tokens\footnote{note that here we use tokens while for the passages and representation of conversation we use characters} for the BERT-C-cq model and 384 for the BERT-C-P-cq model, learning rate of $2$e-$5$ and $3$ training epochs. 

We retrieved at most $1000$ initial candidate clarifications for each passage. 
All experiments were run on a 32GB V100 GPUs. The re-ranking times of $1000$ clarification questions for each conversation took about $1-2$ sec. 
For evaluation metrics we followed the ClariQ leaderboard~\footnote{\UrlFont{https://convai.io}} and used the Recall@30 as the main metrics.

\was{
Details of the setup are given in appendix~\ref{appendix:setup}.
For evaluation metrics, we followed the ClariQ leaderboard\footnote{\UrlFont{https://convai.io}} and used the Recall@30 as the main metric.

For evaluation, we take the conversation context $C^j$ ($j \ge 1$) in the dev/test sets up to the first clarification point, and predict the clarification question at the next utterance $j$. In \textbf{ClariQ}, 
since there is no real conversation context (each topic has an independent set of clarification questions and their answers), it will be just the basic question (i.e., $c_0$), while in \textbf{Support}, $C^j$ can contain several utterances. }

%% file: experiments-results.tex
\subsection{Results}
\label{sec:results}

Table~\ref{results:dev} reports the results 
on the dev sets of the two datasets.\footnote{We compare our methods on the dev sets since in \textbf{Clariq} we had access only to the dev set. We note that in both datasets, the dev sets wer not used during the training, thus they can be regarded as an held-out test set}  On both datasets, each of the BERT re-rankers showed a significant improvement over the initial retrieval 
from the Clarification-questions index (denoted by \textbf{IR-Base}).
For example on \textbf{Support}, \textbf{BERT-C-cq} achieved $R@30$=$0.538$ compared to $R@30$=$0.294$ of \textbf{IR-Base} (an improvement of $82\%$). 

We can further see that the two BERT models (\textbf{BERT-C-cq} and  \textbf{BERT-C-P-cq}), yield quite similar results on both datasets,
but, when fusing their scores (\textbf{BERT-fusion}), there is another improvement of about $2.5\%$ over each of the rankers separately. For example on \textbf{ClariQ}, \textbf{BERT-fusion} achieved $R@30$=$0.791$, compared to $R@30$=$0.77$ of \textbf{BERT-C-cq}.  

\added{This improvement can be attributed to complementary matching that each of the two BERT models learns. The second model learns latent features that are revealed only through the retrieved passages, while the first model works better for cases where the retrieved passages are noisy. For example for query 133 in \textbf{Clariq}, \textit{all men are created equal} (see Figure~\ref{fig:example} above), \textbf{BERT-C-P-cq} could find nine correct clarification questions out of 14 in its top-30 (including those two in the Figure), while \textbf{BERT-C-cq} found only three of them. 
} 


Table~\ref{results:test} shows the official Clariq leaderboard result on the test set. We can see that our method \textbf{BERT-fusion}\footnote{Our run was labeled CogIR in the official leaderboard} was ranked forth but was the second best as a team.
\added{We note that the top performing system (NTES\_ALONG) gave preferences to clarification questions from the test data, capitalizing the specific \textbf{Clariq} properties that test topics came from different domain than the train topics. This is not a valid assumption in general.  In contrast, we treat all clarification questions equally in the given pool of clarification questions.} 

\begin{table}[tbh]
\small
\caption{Retrieval quality on the dev set of the two datasets}
	\label{results:dev}
	\center
\begin{tabular}{|l|c|c|c|c|}
\hline
\textbf{ClariQ} &  \textbf{R@5}      & \textbf{R@10}      & \textbf{R@20}      & \textbf{R@30}      \\ \hline
IR-Base         & .327      & .575      & .669      & .706      \\ \hline
BERT-C-cq       & .352      & .631      & .743      & .770      \\ \hline
BERT-C-P-cq     & .344      & .615      & .750      & .774      \\ \hline
BERT-fusion          & \textbf{.353}      & \textbf{.639}      & \textbf{.758}      & \textbf{.791}      \\ \hline \hline
\textbf{Support}   &           &           &           &           \\ \hline
IR-Base         & .102      & .153      & .269      & .294      \\ \hline
BERT-C-cq       & \textbf{.358}      & \textbf{.410}      & .487      & .538      \\ \hline
BERT-C-P-cq     & .217      & .294      & .487      & .538      \\ \hline
BERT-fusion          & .294      & \textbf{.410}      & \textbf{.500}      & \textbf{.551}      \\ \hline
\end{tabular}
\end{table}

\begin{table}[tbh]
\small
\caption{Retrieval quality on the test set of the ClariQ dataset}
	\label{results:test}
	\center
\begin{tabular}{|l|c|c|c|c|}
\hline
\textbf{ClariQ} &  \textbf{R@5}      & \textbf{R@10}      & \textbf{R@20}      & \textbf{R@30}      \\ \hline
NTES\_ALONG        & .340      & .632      & .833      & .874      \\ \hline
NTES\_ALONG        & .341      & .635      & .831      & .872      \\ \hline
NTES\_ALONG        & .338      & .624      & .817      & .868      \\ \hline
\textbf{BERT-fusion}      & .338      & .631      & .807      & .857      \\ \hline 
TAL-ML       & .339      & .625     & .817      & .856      \\ \hline 
Karl      & .335      & .623     & .799      & .849      \\ \hline 
Soda      & .327      & .606     & .801      & .843      \\ \hline\end{tabular}
\end{table}

%% file: conclusions.tex
\section{Conclusions}
\label{sec:conclusions}

We presented a method for clarification-questions selection in conversational-search scenarios that end with documents as answers. 

We showed that using passages, combined with deep-learning models, improves the quality of the selected clarification questions. 
We evaluated our method on two diversified dataset. 
On both datasets, the usage of passages for clarification-questions re-ranking achieved improvement of $12\%-87\%$ over base IR retrieval. 


%% file: Appendix/passage_retrieval.tex
\subsection{Passage Retrieval details}
\label{appendix:passage}
We use Apache 
Lucene 
for indexing the documents, configured with English language analyzer and default BM25 similarity~\cite{RobertsonBM25}.

\was{For document retrieval, we create a disjunctive query from all words in the conversation $C^j$.  
Following~\cite{agent-assist}, we
treat the dialog query as a verbose query and apply the Fixed-Point (FP) method~\cite{Paik2014} for weighting its words. Yet, compared to ``traditional'' verbose queries, dialogs are further segmented into distinct utterances. Using this observation, we implement an \textit{utterance-biased} extension for enhanced word-weighting. To this end, we first score the various utterances based on the initial FP weights of words they contain. We then propagate utterance scores back to their associated words.}

After retrieving top-k documents, candidate passages are extracted from those documents using a sliding window of fixed size with some overlap. Each retrieved passage $p$ is assigned an initial score based on the coverage of terms in $C^j$ by $p$. The coverage is defined as the sum over all terms in each utterance, using terms' global~ $idf$ (inverse document frequency) and their (scaled) $tf$ (term frequency). 
Let {\small $c$} be a conversation with {\small $n$} utterances {\small $c = u_1, ... u_n$}.  
Passage score is computed as a linear combination of its initial score {\small $score_{init}(p,c)$}
and the score of its enclosing document. Both scores are normalized.

\begin{small}
\begin{equation}
\begin{split}
  score(p,c)& = \lambda * score(d)  + (1 - \lambda) * score_{init}(p,c) \\
\end{split}
\end{equation}
\end{small}
We used lambda=0.5, i.e., fixed equal weights for the document and the passage scores.

The initial passage score {\small $score_{init}(p,c)$} is computed as a weighted sum over its utterances scores {\small $score_{ut}(p, u_i)$}. Utterance scores are discounted such that later utterances have greater effect on the passage score.

\begin{small}
\begin{equation}
\begin{split}
score_{init}(p,c)& = \sum_{i=1}^n ~ weight_{ut}(i) * score_{ut}(p,u_i) \\
weight_{ut}(i)& = discount\_factor^{(n-i)} \\
discount\_factor& = 0.85 \\
\end{split}
\end{equation}
\end{small}

Utterance score {\small $score_{ut}(p,u)$} reflects utterance's terms coverage by the passage, considering terms' global~ {\small $idf$} (inverse document frequency) and their (scaled) {\small $tf$} (term frequency).
Multiple coverage scorers are applied, which differ by their term frequency scaling schemes.
Finally, the utterance score is a product of these coverage scores {\small $score_{cov}(p,u)$}.

\begin{small}
\begin{equation}
\begin{split}
score_{ut}(p,u)& = \Pi_{j=1}^m ~score_{cov_j}(p,u) \\
m& = 2 \textnormal{~~~(two scaling schemes are employed)} \\
score_{cov_j}(p,u)& = \sum_{t \in t^{pu}} idf(t) * scale_j(t, p) \\
t^{pu}& = t^u \bigcap t^p \textnormal{~~~(terms appearing in both)} \\
t^p, t^u& = \textnormal{~~~(passage terms, utterance terms)} \\
\end{split}
\end{equation}
\end{small}

Different scaling schemes provide different interpretations of terms' importance. We combine two {\small $tf$} scaling methods, one that scales by a BM25 term score, and another that scales by the minimum of {\small $tf(t)$} in the utterance and passage.

\begin{small}
\begin{equation}
\begin{split}
  scale_1& = BM25(t, p) \\
  scale_2& = min(tf(t,p), tf(t,c)) \\
\end{split}
\end{equation}
\end{small}

The final passage score is a linear combinations of its initial score and the score of the document it is extracted from.
Candidate passage ranking exploits a cascade of scorers.

%% file: main.bbl
\begin{thebibliography}{16}
\expandafter\ifx\csname natexlab\endcsname\relax\def\natexlab#1{#1}\fi

\bibitem[{Aliannejadi et~al.(2020)Aliannejadi, Kiseleva, Chuklin, Dalton, and
  Burtsev}]{aliannejadi2020convai3}
Mohammad Aliannejadi, Julia Kiseleva, Aleksandr Chuklin, Jeff Dalton, and
  Mikhail Burtsev. 2020.
\newblock \href {http://arxiv.org/abs/2009.11352} {Convai3: Generating
  clarifying questions for open-domain dialogue systems (clariq)}.

\bibitem[{Aliannejadi et~al.(2019)Aliannejadi, Zamani, Crestani, and
  Croft}]{Aliannejadi-sigir19}
Mohammad Aliannejadi, Hamed Zamani, Fabio Crestani, and W.~Bruce Croft. 2019.
\newblock \href {https://doi.org/10.1145/3331184.3331265} {Asking clarifying
  questions in open-domain information-seeking conversations}.
\newblock In \emph{Proceedings of the 42nd International ACM SIGIR Conference
  on Research and Development in Information Retrieval}, SIGIR'19, page
  475–484, New York, NY, USA. Association for Computing Machinery.

\bibitem[{Amitay et~al.(2005)Amitay, Darlow, Konopnicki, and
  Weiss}]{amitayQanchors}
Einat Amitay, Adam Darlow, David Konopnicki, and Uri Weiss. 2005.
\newblock Queries as anchors: selection by association.
\newblock In \emph{Proceedings of the 16th ACM Conference on Hypertext and
  Hypermedia}, pages 193--201.

\bibitem[{Devlin et~al.(2019)Devlin, Chang, Lee, and Toutanova}]{BERT}
Jacob Devlin, Ming-Wei Chang, Kenton Lee, and Kristina Toutanova. 2019.
\newblock \href {https://doi.org/10.18653/v1/N19-1423} {{BERT}: Pre-training of
  deep bidirectional transformers for language understanding}.
\newblock In \emph{Proceedings of the 2019 Conference of the North {A}merican
  Chapter of the Association for Computational Linguistics: Human Language
  Technologies, Volume 1 (Long and Short Papers)}, pages 4171--4186,
  Minneapolis, Minnesota. Association for Computational Linguistics.

\bibitem[{Ganhotra et~al.(2020)Ganhotra, Roitman, Cohen, Mills, Gunasekara,
  Mass, Joshi, Lastras, and Konopnicki}]{agent-assist}
Jatin Ganhotra, Haggai Roitman, Doron Cohen, Nathaniel Mills, R.~Chulaka
  Gunasekara, Yosi Mass, Sachindra Joshi, Luis~A. Lastras, and David
  Konopnicki. 2020.
\newblock \href {https://www.aclweb.org/anthology/2020.emnlp-main.25/}
  {Conversational document prediction to assist customer care agents}.
\newblock In \emph{Proceedings of the 2020 Conference on Empirical Methods in
  Natural Language Processing, {EMNLP} 2020, Online, November 16-20, 2020},
  pages 349--356. Association for Computational Linguistics.

\bibitem[{Hoffer and Ailon(2015)}]{hoffer_ailon2015}
Elad Hoffer and Nir Ailon. 2015.
\newblock \href {http://arxiv.org/abs/1412.6622} {Deep metric learning using
  triplet network}.
\newblock In \emph{3rd International Conference on Learning Representations,
  {ICLR} 2015, San Diego, CA, USA, May 7-9, 2015, Workshop Track Proceedings}.

\bibitem[{Joshi et~al.(2018)Joshi, Peters, and Hopkins}]{joshi2018extending}
Vidur Joshi, Matthew Peters, and Mark Hopkins. 2018.
\newblock \href {http://arxiv.org/abs/1805.06556} {Extending a parser to
  distant domains using a few dozen partially annotated examples}.

\bibitem[{Krasakis et~al.(2020)Krasakis, Aliannejadi, Voskarides, and
  Kanoulas}]{Krasakis_2020}
Antonios~Minas Krasakis, Mohammad Aliannejadi, Nikos Voskarides, and Evangelos
  Kanoulas. 2020.
\newblock \href {https://doi.org/10.1145/3409256.3409817} {Analysing the effect
  of clarifying questions on document ranking in conversational search}.
\newblock \emph{Proceedings of the 2020 ACM SIGIR on International Conference
  on Theory of Information Retrieval}.

\bibitem[{Liu et~al.(2019)Liu, Ott, Goyal, Du, Joshi, Chen, Levy, Lewis,
  Zettlemoyer, and Stoyanov}]{Roberta}
Yinhan Liu, Myle Ott, Naman Goyal, Jingfei Du, Mandar Joshi, Danqi Chen, Omer
  Levy, Mike Lewis, Luke Zettlemoyer, and Veselin Stoyanov. 2019.
\newblock \href {http://arxiv.org/abs/1907.11692} {Roberta: {A} robustly
  optimized {BERT} pretraining approach}.
\newblock \emph{CoRR}, abs/1907.11692.

\bibitem[{Mass et~al.(2019)Mass, Roitman, Erera, Rivlin, Weiner, and
  Konopnicki}]{mass2019study}
Yosi Mass, Haggai Roitman, Shai Erera, Or~Rivlin, Bar Weiner, and David
  Konopnicki. 2019.
\newblock \href {http://arxiv.org/abs/1908.06780} {A study of bert for
  non-factoid question-answering under passage length constraints}.
\newblock \emph{CoRR}, abs/1908.06780.

\bibitem[{Ou and Lin(2020)}]{NTES-ALONG}
Wenjie Ou and Yue Lin. 2020.
\newblock \href {http://arxiv.org/abs/2010.14202} {A clarifying question
  selection system from ntes\_along in convai3 challenge}.
\newblock \emph{CoRR}, abs/2010.14202.

\bibitem[{Paik and Oard(2014)}]{Paik2014}
Jiaul~H. Paik and Douglas~W. Oard. 2014.
\newblock \href {https://doi.org/10.1145/2661829.2661957} {A fixed-point method
  for weighting terms in verbose informational queries}.
\newblock In \emph{Proceedings of the 23rd ACM International Conference on
  Conference on Information and Knowledge Management}, CIKM ’14, page
  131–140, New York, NY, USA. Association for Computing Machinery.

\bibitem[{Rao and Daum{\'e}~III(2018)}]{rao-daume-acl-2018}
Sudha Rao and Hal Daum{\'e}~III. 2018.
\newblock \href {https://doi.org/10.18653/v1/P18-1255} {Learning to ask good
  questions: Ranking clarification questions using neural expected value of
  perfect information}.
\newblock In \emph{Proceedings of the 56th Annual Meeting of the Association
  for Computational Linguistics (Volume 1: Long Papers)}, pages 2737--2746,
  Melbourne, Australia. Association for Computational Linguistics.

\bibitem[{Robertson and Zaragoza(2009)}]{RobertsonBM25}
Stephen Robertson and Hugo Zaragoza. 2009.
\newblock \href {https://doi.org/10.1561/1500000019} {The probabilistic
  relevance framework: Bm25 and beyond}.
\newblock \emph{Found. Trends Inf. Retr.}, 3(4):333--–389.

\bibitem[{Wu(2012)}]{combsum}
Shengli Wu. 2012.
\newblock \href {https://doi.org/10.1007/978-3-642-28866-1} {\emph{Data Fusion
  in Information Retrieval}}, volume~13.

\bibitem[{Zamani et~al.(2020)Zamani, Dumais, Craswell, Bennett, and
  Lueck}]{zamani20}
Hamed Zamani, Susan Dumais, Nick Craswell, Paul Bennett, and Gord Lueck. 2020.
\newblock \href {https://doi.org/10.1145/3366423.3380126} {Generating
  clarifying questions for information retrieval}.
\newblock In \emph{Proceedings of The Web Conference 2020}, WWW '20, page
  418–428, New York, NY, USA. Association for Computing Machinery.

\end{thebibliography}
